# Fine-tuned Pre-trained Mask R-CNN Models for Surface Object Detection

Haruhiro Fujita
*Faculty of Management and Information Sciences*
*Niigata University of International and Information Studies* Niigata City, Japan
fujita@nuis.ac.jp

Masatoshi Itagaki
*BSN iNet Co. Ltd.*
Niigata City, Japan
masa-ita@bsnnet.co.jp

Kenta Ichikawa
*BSN iNet Co. Ltd.*
Niigata City, Japan
ichikawa@bsnnet.co.jp

Yew Kwang Hooi
*Department of Computer and Information Sciences*
*Universiti Teknologi PETRONAS*
Bandar Sri Iskandar, Malaysia
yewkwanghooi@utp.edu.my

Kazutaka Kawano
*Department of Curatorial Research*
*Tokyo National Museum*
Tokyo, Japan
kawano4191kka489@gmail.com

Ryo Yamamoto
*Department of Cutorial Planning*
*Tokyo National Museum*
Tokyo, Japan
yamamoto-r62@nich.go.jp

***Abstract*—This study evaluates road surface object detection tasks using four Mask R-CNN models as a pre-study of surface deterioration detection of stone-made archaeological objects. The models were pre-trained and fine-tuned by COCO datasets and 15,188 segmented road surface annotation tags. The quality of the models were measured using Average Precisions and Average Recalls. Result indicates substantial number of counts of false negatives, i.e. "left detection" and "unclassified detections". A modified confusion matrix model to avoid prioritizing IoU is tested and there are notable true positive increases in bounding box detection, but almost no changes in segmentation masks.**

## I. INTRODUCTION

Computer vision is a broad science field handling various types of cognitive image processing including object detection [1]. Object detection is a task of finding different objects in an image and classifying them. Object detection is a critical task in autonomous driving where advanced deep learning models in conjunction with many AI sensors play major roles to control the system [2]. Various objects on the road surface including potholes and severe cracks require vehicle to identify and maneuver over them smoothly in real-time.

R-CNN (Region Convolutional Neural Network) is a very effective method for identifying objects. Results are shown as bounding box around detected objects with suitable class label. Ever since introduction of R-CNN for object detection in 2014, various improved algorithms have been introduced, such as Fast R-CNN, Faster R-CNN and Mask R-CNN.

R-CNN extracts region of proposals, compute the CNN features and classify the regions accordingly. Selective search uses windows of various dimensions to determine regions based on pixel features such as color, intensity and texture. The identified regions are warped into a standard square size and parsed to a Convolutional Neural Network for features extraction. Finally, Support Vector is used to classify the regions.

Although speed has improved through optimization of the learning process, speed remains a critical problem for real-time object detection for surface object detection from a moving sensor. Various existing CNN models can be studied for improving bounding box potentials for surface object detection.

This manuscript is organized as follows:- literature review of surface object detection technologies and applications for road surface object detections; specific objectives of this study; methods of study; result of the testing the models; discussion of the profiles of each model and pinpointing critical areas for improvement.

## II. LITERATURE REVIEW

### A. Superpixel and Instance Segmentation

Superpixel is used as the base for instance segmentation, as a graphical preparatory clustering method. It clusters pixels of vicinity in terms of geometric and color spaces prior to object segmentation using SLIC (Simple Linear Iterative Clustering) algorithm [3, 4]. This study uses superpixel as the base for instance segmentation. It is defined and enhanced by the Common Object in Context (COCO). COCO is a large image-based datasets with object segmentation, recognition in context, differentiation of objects (cars, people, houses) and stuffs (roads, sky, wall), 200k labeled images, 1.5 million object instances, 80 object categories, 91 stuff categories, and 5 captions in each image [5].

Instance segmentation detects an object under overlap of other objects precisely. However, it does not label all of the pixels of the image, as it segments only the region of interests [6]. The state-of-the-art instance segmentation approach is the detection-based method that predicts the mask for each region after acquiring the instance region [7].

### B. Mask Region Convolutional Neural Network (Mask R-CNN)

Fast R-CNN, introduced in 2015 to improve the speed of R-CNN, is a complex pipeline. It requires huge number of forward passes and separate training of three models for image features generation, classification and regression respectively. Faster R-CNN improves Fast R-CNN for less complex training through ROI Pooling, a technique that limits the number of unnecessary forward passes through shared computation result. However, Faster R-CNN is limited to detect object in bounding boxes only.

Enters Mask Region Convolutional Neural Network (Mask R-CNN), as proposed by [8]. Mask R-CNN is an extended model of the Faster R-CNN [9] which provides better detection of objects in segmentations. Mask R-CNN adds learning segmentation masks (called Mask Branch) sub-module onto Faster R-CNN. The sub-module predicts segmentation masks on each Region of Interest (RoI) [6] in parallel with each other using convolution arrays of data for classification and bounding box regression. The mask module is a small fully convolutional network (FCN) applied to each RoI, predicting a segmentation mask for each pixel.

### C. *Four Mask R-CNN models in TensorFlow Object Detection API*

TensorFlow Object Detection API is an open archive of deep learning models for object detection. It contains pre-trained models with multiple datasets including COCO. The pre-trained models generate bounding boxes around objects. From the various models, four Mask R-CNN models which generate masks (instance segments) of identified objects are selected for this study [10].

### D. *Road damage detection system and models*

An application of object detection is road damage detection. The first reported road surface object detection models using Mask R-CNN are cited in [11], as shown Fig. 1.

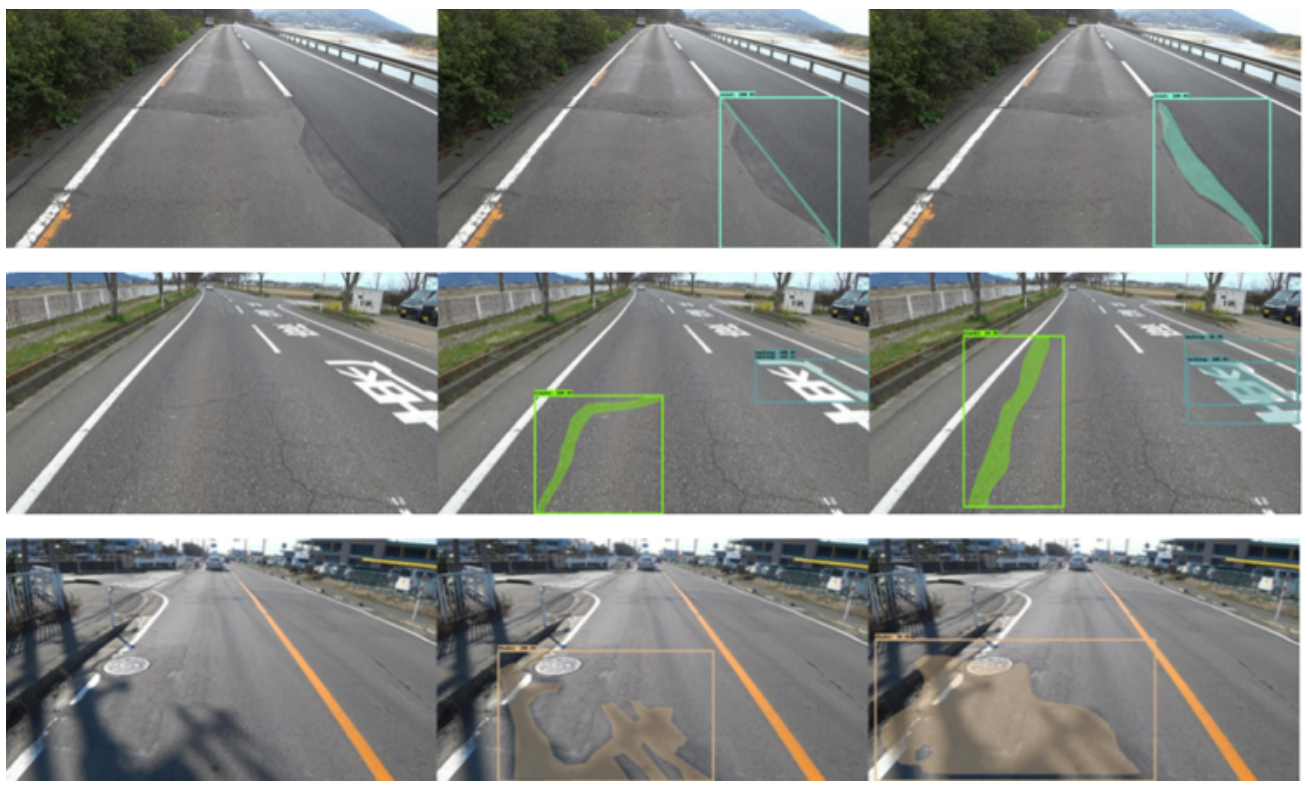

Fig. 1. Road surface object detection using one of Mask R-CNN models

Another related work is that of Maeda et al.[12] who had trained Single Shot Detection (SSD) models on their dataset captured with Smartphones.

## III. OBJECTIVES

The objectives of this study are to:-

- evaluate several pre-trained Mask R-CNN models in TensorFlow Object Detection API
- fine-tune the models by annotation data of road objects
- study the applicability of the models to the road object detection

## IV. METHODOLOGY

The methodologies are the same as previously reported in [11]. To optimize performance of road surface detection, an enhanced method was applied to train pre-learned four Mask R-CNN models in TensorFlow Object Detection API. The models are then evaluated using sorted annotation test data.

### A. *Data and Model Preparation*

Refer to TABLE I. The annotation data contains 12 classes of road damages and cognitive objects in the road images to differentiate those damages from others. Microsoft Visual Object Tagging Tool (VoTT) was used for annotating 1,418 road color images with a size of 3840 x 2160. All visible objects of the predefined classes in two third bottom part of each image were segmented and tagged.

TABLE I. ROAD OBJECT ANNOTATION DATA OBTAINED FROM 1418 ROAD IMAGES FOR MODEL FINETUNING

| | | Datasets/segmentations | | | |
|---|---|---|---|---|---|
| classes | tag | training | validation | testing | total |
| Linear Cracks | Crack1 | 2,315 | 468 | 455 | **3,238** |
| Grid Cracks | Crack2 | 524 | 103 | 101 | **728** |
| Pavement Joints | Joint | 941 | 198 | 219 | **1,358** |
| Patchings | Patching | 377 | 102 | 77 | **556** |
| Fillings | Filling | 1,240 | 207 | 187 | **1,634** |
| Pot-holes | Pothole | 143 | 39 | 14 | **196** |
| Manholes | Manhole | 296 | 61 | 52 | **409** |
| Stains | Stain | 98 | 22 | 12 | **132** |
| Shadow | Shadow | 1,084 | 252 | 212 | **1,548** |
| Pavement Markings | Marking | 1,239 | 241 | 297 | **1,777** |
| Scratches on Markings | Scratch | 2,433 | 494 | 576 | **3,503** |
| Grid Crack in Patchings | Patching2 | 73 | 19 | 17 | **109** |
| **Total** | | **10,763** | **2,206** | **2,219** | **15,188** |

The most numerous segments were “Scratches on Markings” class at total number of 3,503 segments. This is followed by “Linear Cracks” at 3,238 segments. The minimum segments were “Grid Cracks in Patchings” at 109 segments followed by “Stains” and “Pot-holes”. The segments of each class were divided into datasets for training, training validation (validation) and testing at ratio of 0.7:0.15:0.15.

### B. *Computation Resources*

The tests were conducted on a machine with Intel Core i7-8700K (6 cores/12 threads/3.70GHz) CPU, 32GB CPU memory, Nvidia GeForce GTX 1080 GPU with 2560 CUDA core, GPU clocks of 1607/1733 MHz, 8GB GDDRSX GPU memory and 320GB/s memory bandwidth. The language for programming is Python 3.6.9 using libraries from Object Detection API version 1 on top of TensorFlow 1.15.0.

### C. *Methods*

Main steps of the methods are performing annotation, mask data making; model training and fine tuning.

#### 1) *Annotation and Mask Data Making:*

Refer to Fig. 2. The files output by the Visual Object Tagging Tool (VoTT) were converted to intermediate annotation files. Next, the files were converted into TensorFlow Record files for training, validation and testing. The conversions were done using three separate conversion script executive programs in Python. The last conversion script (vott2tfrecords.py) modified the original image size of 3840x2160 (width x height) to match with the training image size of 960x540 used in machine learning.

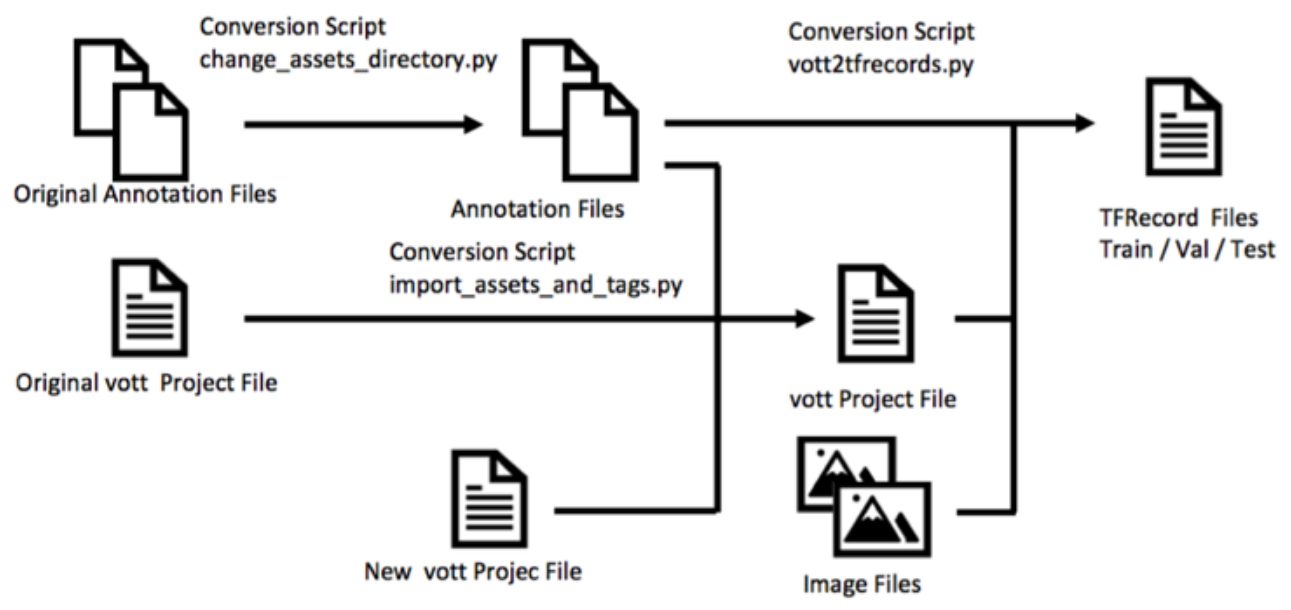

Fig. 2. Flow for annotation and data analysis.

*2) Model Training and Fine Tuning:*

Refer to Fig. 3. Pre-trained model checkpoint file, TensorFlow Record files, ground truth annotation data/labels ("tf_label_map.pbtxt") and pipeline.config are used for training of object detection. The pretrained model checkpoint is based on MS COCO dataset. The TensorFlow Record files contain both datasets for training and validation. The training generated Fine Tuned Checkpoint file fromwhich an object detection model was acquired. Finally, the fine-tuned detection model was evaluated using validation and testing datasets. The inference result produces precision metric, recall metric and confusion matrix. The training was done for each four Mask R-CNN models.

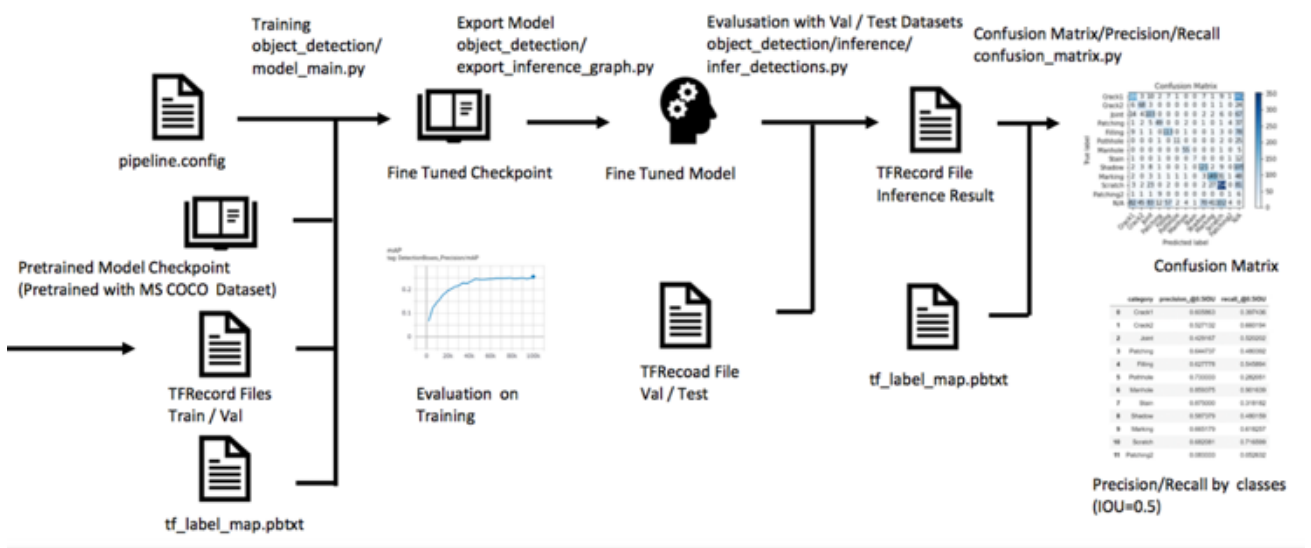

Fig. 3. Training and evaluation processes of pre-trained Mask R-CNN models.

## D. Model Evaluation Metrics

Various metrics used for model evaluation of this work are mean average precisions (mAP) and average recalls (AR) at various levels of Intersection on Unit(IoU). The Mask R-CNN has a region proposal network layer that makes multiple inferences simultaneously on the class classification, the segmentation and the mask areas resulting six loss metrics.

Besides above model-wise metrics, the average precisions and the average recalls at IoU=0.5 are used for all twelve road object classes.

# V. RESULTS

## A. Evaluation Metrics of Mask R-CNN variants

*1) Inceptions v2 COCO:*

TABLE II and TABLE III shows various metrics of the model on bounding boxes and segmentation masks respectively. The values of mAP(IoU=.50:.05:.95), mAP(IoU=.50) and mAP(IoU=.75) are 0.2432, 0.4382 and 0.2482 on bounding boxes, and 0.1600, 0.3257 and 0.1279 respectively, with notable reduction in those metrics on segmentation masks. The Precision mAP(small) for small objects = 0.0365 and 0.0133 on bounding boxes and segmentation masks respectively are notably small, compared with those Precision mAP(large) of large objects and medium objects. The Average Recall for small, medium and large object are 0.1166, 0.3132 and 0.4717 on boundary boxes and 0.1021, 0.2528 and 0.2732 on segmentation masks respectively.

The detection precisions at IoU=.50 found in our target damage classes of linear cracks (Crack1), grid cracks (Crack2), pot-holes, scratches on markings, grid cracks in patchings are 0.4085, 0.4958, 0.5714, 0.5934 and 0.4000 respectively..

TABLE II. MODEL METRICS OF MASK R-CNN INCEPTION V2 COCO ON BOUDING BOX

Precision | Recall

| tag | precision_@0.5IOU | recall_@0.5IOU |
|---|---|---|
| Crack1 | 0.4085 | 0.2549 |
| Crack2 | 0.4958 | 0.5842 |
| Joint | 0.3602 | 0.3881 |
| Patching | 0.6346 | 0.4286 |
| Filling | 0.4773 | 0.3369 |
| Pothhole | 0.5714 | 0.2857 |
| Manhole | 0.8298 | 0.7500 |
| Stain | 0.0000 | 0.0000 |
| Shadow | 0.3975 | 0.3019 |
| Marking | 0.6224 | 0.6162 |
| Scratch | 0.5934 | 0.6233 |
| Patching2 | 0.4000 | 0.1176 |

Mask RCNN Inception v2 coco

| Index | |
|---|---|
| Precision mAP | 0.2432 |
| Precision mAP@.50IOU | 0.4382 |
| Precision mAP@.75IOU | 0.2482 |
| Precision mAP (large) | 0.2953 |
| Precision mAP (medium) | 0.1711 |
| Precision mAP (small) | 0.0365 |
| Recall AR@1 | 0.1995 |
| Recall AR@10 | 0.3765 |
| Recall AR@100 | 0.4140 |
| Recall AR@100 (large) | 0.4710 |
| Recall AR@100 (medium) | 0.3132 |
| Recall AR@100 (small) | 0.1166 |
| BoxClassifierLoss classification loss | 0.5757 |
| BoxClassifierLoss localization loss | 0.4821 |
| BoxClassifierLoss mask loss | 0.9846 |
| RPNLoss localization loss | 1.3293 |
| RPNLoss objectness loss | 0.3359 |
| Total Loss | 3.7076 |

TABLE III. MODEL METRICS OF MASK R-CNN INCEPTION V2 COCO ON SEGMENTATION MASK

Precision | Recall

| tag | precision_@0.5IOU | recall_@0.5IOU |
|---|---|---|
| Crack1 | 0.2285 | 0.1341 |
| Crack2 | 0.4622 | 0.5446 |
| Joint | 0.1200 | 0.1233 |
| Patching | 0.6863 | 0.4545 |
| Filling | 0.1742 | 0.1230 |
| Pothhole | 0.5714 | 0.2857 |
| Manhole | 0.8298 | 0.7500 |
| Stain | 0.0000 | 0.0000 |
| Shadow | 0.1987 | 0.1415 |
| Marking | 0.3767 | 0.3704 |
| Scratch | 0.3565 | 0.3646 |
| Patching2 | 0.6000 | 0.1765 |

Mask RCNN Inception v2 coco

| Index | |
|---|---|
| Precision mAP | 0.1600 |
| Precision mAP@.50IOU | 0.3257 |
| Precision mAP@.75IOU | 0.1279 |
| Precision mAP (large) | 0.2012 |
| Precision mAP (medium) | 0.1210 |
| Precision mAP (small) | 0.0133 |
| Recall AR@1 | 0.1559 |
| Recall AR@10 | 0.2533 |
| Recall AR@100 | 0.2713 |
| Recall AR@100 (large) | 0.2732 |
| Recall AR@100 (medium) | 0.2528 |
| Recall AR@100 (small) | 0.1021 |
| BoxClassifierLoss classification loss | 0.5757 |
| BoxClassifierLoss localization loss | 0.4821 |
| BoxClassifierLoss mask loss | 0.9846 |
| RPNLoss localization loss | 1.3248 |
| RPNLoss objectness loss | 0.3363 |
| Total Loss | 3.7035 |

Fig. 4 shows the model's confusion matrix of the model on bounding boxes (left) and segmentation masks (right) respectively. The horizontal rows represent the true label class (or the ground truth class). The vertical column represents the predicted class. Each cell contains count number for the corresponding true class and predicted class.

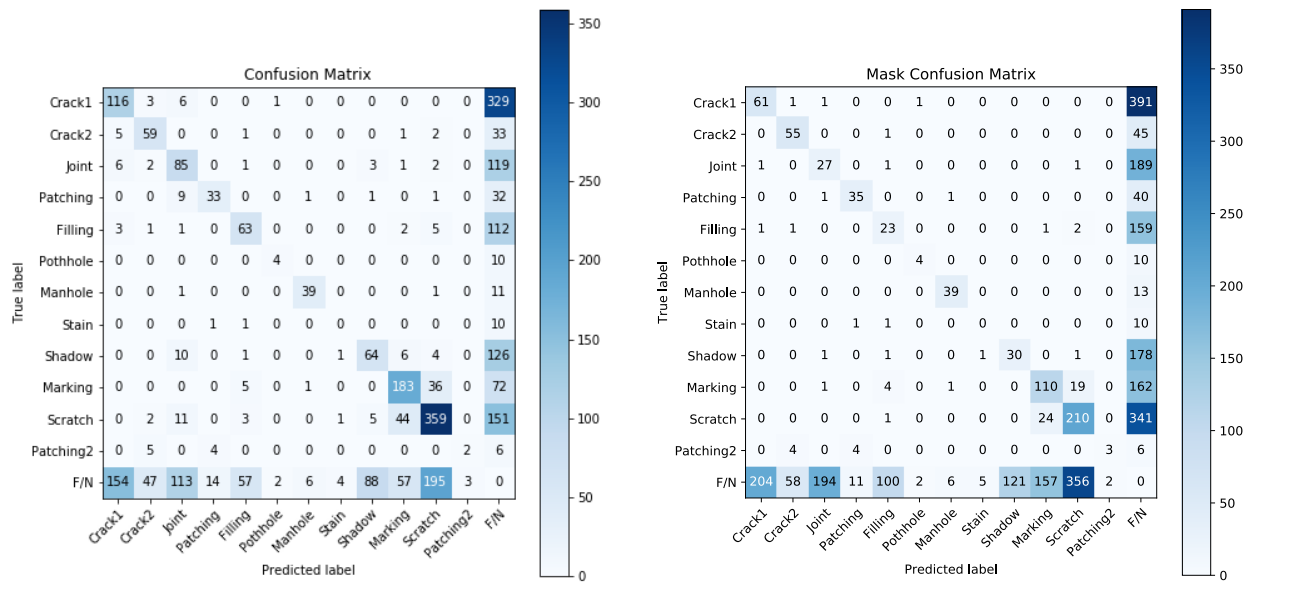

Fig. 4. *Confusion Matrix of Mask RCNN Inception v2 COCO ON BOUDING BOX (LEFT) AND SEGMENTATION MASK (RIGHT)*

For example, in the left matrix of the bounding boxes, in the first horizontal row from the left, 116 is the count of true class of "Crack1" and predicted class of "Crack1" or linear cracks. It indicates the number of boundary boxes of true class linear cracks predicted correctly. Other cells in the same row indicates "Crack1" wrongly predicted as other classes.

The last count 329 is the number of boundary boxes where ground truths of linear crack existed but is not detected in any classes. This means "the model could not predict ground truths as any classes", shortly "left detection" or False Negative (F/N). The bottom row are the "False Negative" counts, i.e. the boundary boxes that were detected but not classified at any classes for lower IoU values. Vice versa in the confusion matrix of segmentation masks.

Substantial "left detection" counts are found in linear cracks, grid cracks, joints, patchings, fillings, potholes, stains, shadows, patchings with grid cracks, where the counts exceed 50% of the correct prediction counts, both in bounding boxes and segmentation masks. Substantial "unclassified detections" (where counts exceed 50% of True Positive) are found in linear cracks, grid cracks, joints, fillings, potholes, shadows and fillings with grid cracks, both in bounding boxes and segmentation masks.

### 2) *Inception ResNet v2 Atrous COCO:*

TABLE IV and TABLE V show mAP(IoU=.50:.05:.95), mAP(IoU=.50) and mAP(IoU=.75) of the model, which are 0.2756, 0.4678 and 0.2815 on bounding boxes, and 0.2096, 0.4038 and 0.1982 on segmentation masks respectively. The mAP for small objects Precision mAPs (small) = 0.0776 and 0.0330 are notably small as compared with large objects Precision mAPs (large) = 0.3393 and 0.2659, and medium objects Precision mAPs (medium) = 0.1853 and 0.1552 in bounding boxes and segmentation masks respectively. In comparison, there are notable reductions in all precision metrics on segmentation masks. The Average Recall for small, medium and large object are 0.1552, 0.3562 and 0.4705 on boundary boxes and 0.1314, 0.3106 and 0.3325 on segmentation masks respectively. The detection precisions at IoU=.50 found in our target damage classes of linear cracks, grid cracks, potholes, scratches and markings are 0.3791, 0.5780, 0.4328, 0.6391 and 0.6268 on boundary boxes and 0.3571, 0.5321, 0.5455, 0.5937 and 0.5942 on segmentation masks respectively. Smaller precisions are found in grid cracks in patchings of 0.3000 and 0.2000 on boundary boxes and segmentation masks respectively, compared with those of Mask R-CNN Inception v2 COCO.

TABLE IV. MODEL METRICS OF MASK R-CNN INCEPTION RESNET V2 ATROUS COCO ON BOUNDING BOXES

Precision | Recall

| tag | precision_@0.5IOU | recall_@0.5IOU |
|---|---|---|
| Crack1 | 0.3791 | 0.3275 |
| Crack2 | 0.5780 | 0.6238 |
| Joint | 0.4328 | 0.3973 |
| Patching | 0.5781 | 0.4805 |
| Filling | 0.5000 | 0.4171 |
| Pothhole | 0.5455 | 0.4286 |
| Manhole | 0.9167 | 0.8462 |
| Stain | 0.0909 | 0.0833 |
| Shadow | 0.4870 | 0.3538 |
| Marking | 0.6268 | 0.5825 |
| Scratch | 0.6391 | 0.6701 |
| Patching2 | 0.3000 | 0.1765 |

Mask RCNN Inception ResNet v2 coco

| Index | |
|---|---|
| Precision mAP | 0.2756 |
| Precision mAP@.50IOU | 0.4678 |
| Precision mAP@.75IOU | 0.2815 |
| Precision mAP (large) | 0.3393 |
| Precision mAP (medium) | 0.1853 |
| Precision mAP (small) | 0.0776 |
| Recall AR@1 | 0.2401 |
| Recall AR@10 | 0.4215 |
| Recall AR@100 | 0.4368 |
| Recall AR@100 (large) | 0.4705 |
| Recall AR@100 (medium) | 0.3562 |
| Recall AR@100 (small) | 0.1552 |
| BoxClassifierLoss classification loss | 0.6616 |
| BoxClassifierLoss localization loss | 0.4585 |
| BoxClassifierLoss mask loss | 1.0166 |
| RPNLoss localization loss | 1.4781 |
| RPNLoss objectness loss | 0.4526 |
| Total Loss | 4.0673 |

TABLE V. MODEL METRICS OF MASK R-CNN INCEPTION RESNET V2 ATROUS COCO ON SEGMENTATION MASKS

Precision | Recall

| tag | precision_@0.5IOU | recall_@0.5IOU |
|---|---|---|
| Crack1 | 0.3571 | 0.3077 |
| Crack2 | 0.5321 | 0.5743 |
| Joint | 0.2814 | 0.2557 |
| Patching | 0.5938 | 0.4935 |
| Filling | 0.3910 | 0.3262 |
| Pothhole | 0.5455 | 0.4286 |
| Manhole | 0.9167 | 0.8462 |
| Stain | 0.0909 | 0.0833 |
| Shadow | 0.4133 | 0.2925 |
| Marking | 0.5942 | 0.5522 |
| Scratch | 0.5937 | 0.6215 |
| Patching2 | 0.2000 | 0.1176 |

Mask RCNN Inception ResNet v2 coco

| Index | |
|---|---|
| Precision mAP | 0.2096 |
| Precision mAP@.50IOU | 0.4038 |
| Precision mAP@.75IOU | 0.1982 |
| Precision mAP (large) | 0.2659 |
| Precision mAP (medium) | 0.1552 |
| Precision mAP (small) | 0.0330 |
| Recall AR@1 | 0.2001 |
| Recall AR@10 | 0.3212 |
| Recall AR@100 | 0.3296 |
| Recall AR@100 (large) | 0.3325 |
| Recall AR@100 (medium) | 0.3106 |
| Recall AR@100 (small) | 0.1314 |
| BoxClassifierLoss classification loss | 0.6616 |
| BoxClassifierLoss localization loss | 0.4585 |
| BoxClassifierLoss mask loss | 1.0168 |
| RPNLoss localization loss | 1.4782 |
| RPNLoss objectness loss | 0.4505 |
| Total Loss | 4.0655 |

See Fig. 5 confusion matrices on bounding boxes (left) and segmentation masks (right). They indicate substantial "left detection" (which exceeds 50% of True Positive) in all classes, except for grid cracks (bounding boxes only), manholes, markings (bounding boxes only) and scratches (bounding boxes only). The substantial "unclassified detections" counts (which exceeds 50% of True Positive) are found in linear cracks, grid cracks, joints, fillings, stains, shadows and fillings with grid cracks.

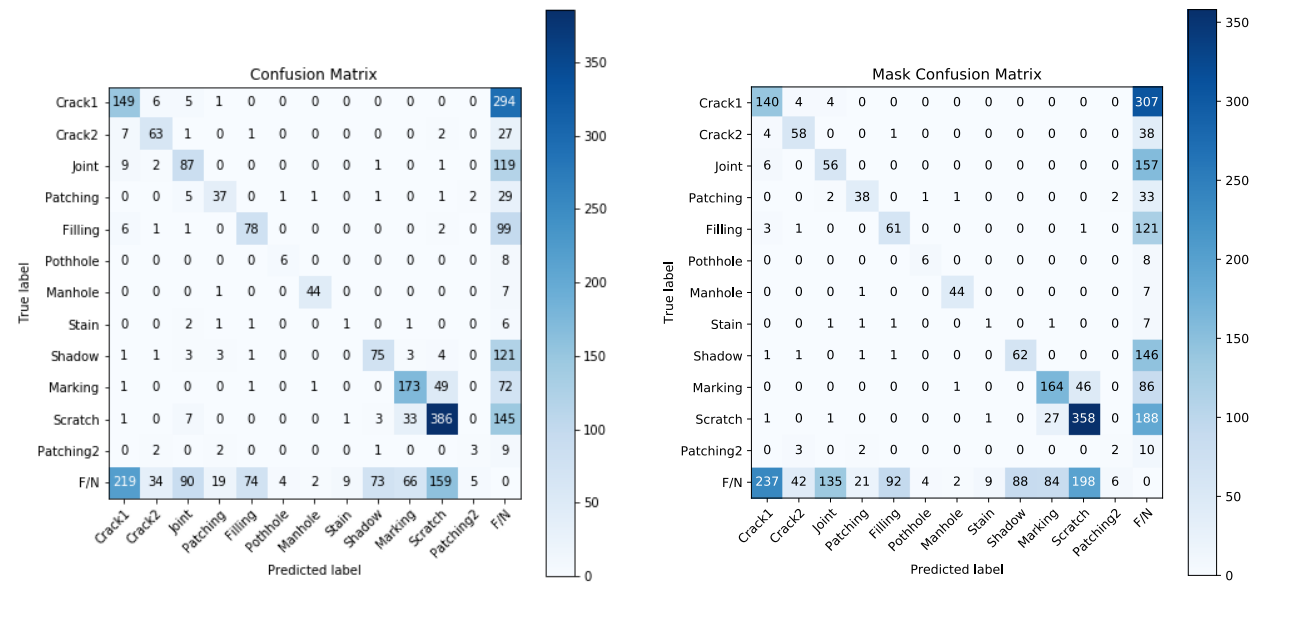

Fig. 5. *Confusion Matrix of Mask R-CNN Inception ResNet v2 atrous COCO on bounding boxes (left) and segmentation masks (right)*

### 3) *Inception ResNet50 Atrous COCO:*

TABLE VI and TABLL VII show Inception RestNet50 Atrous COCO's mAP(IoU=.50:.05:.95), mAP(IoU=.50) and

mAP(IoU=.75) are 0.2810, 0.4789 and 0.2965 on bounding boxes, and 0.2104, 0.4140 and 0.1934 on segmentation masks respectively. The mAPs for small objects of 0.0800 and 0.0452 on bounding boxes and segmentation masks are more than twice and three times as much as the same metric of Mask R-CNN Inception v2 COCO. Notable reductions in metrics on segmentation masks are found both in mAPs and ARs, however their magnitudes are much smaller in ARs. A notable smaller detection precisions at IoU=.50 on segmentation masks are found in those classes of linear cracks, joints, fillings and stains, which are 0.2549, 0.2653, 0.2687 and 0.2222 respectively.

TABLE VI. MODEL METRICS OF MASK R-CNN RESNET50 ATROUS COCO ON BOUNDING BOXES

Precision | Recall

| tag | precision_@0.5IOU | recall_@0.5IOU |
|---|---|---|
| Crack1 | 0.3050 | 0.3077 |
| Crack2 | 0.4024 | 0.6733 |
| Joint | 0.4286 | 0.3836 |
| Patching | 0.6066 | 0.4805 |
| Filling | 0.4229 | 0.4545 |
| Pothhole | 0.5385 | 0.5000 |
| Manhole | 0.9756 | 0.7692 |
| Stain | 0.2222 | 0.1667 |
| Shadow | 0.4424 | 0.4528 |
| Marking | 0.6232 | 0.5960 |
| Scratch | 0.6509 | 0.6701 |
| Patching2 | 0.5000 | 0.2941 |

Mask RCNN ResNet 50 Atrous coco

| Index | |
|---|---|
| Precision mAP | 0.2810 |
| Precision mAP@.50IOU | 0.4789 |
| Precision mAP@.75IOU | 0.2965 |
| Precision mAP (large) | 0.3257 |
| Precision mAP (medium) | 0.2063 |
| Precision mAP (small) | 0.0800 |
| Recall AR@1 | 0.2282 |
| Recall AR@10 | 0.4147 |
| Recall AR@100 | 0.4599 |
| Recall AR@100 (large) | 0.5064 |
| Recall AR@100 (medium) | 0.3736 |
| Recall AR@100 (small) | 0.1442 |
| BoxClassifierLoss classification loss | 0.6451 |
| BoxClassifierLoss localization loss | 0.4884 |
| BoxClassifierLoss mask loss | 0.9162 |
| RPNLoss localization loss | 1.5226 |
| RPNLoss objectness loss | 0.3437 |
| Total Loss | 3.9161 |

TABLE VII. MODEL METRICS OF MASK R-CNN RESNET50 ATROUS COCO ON SEGMENTATION MASKS

Precision | Recall

| tag | precision_@0.5IOU | recall_@0.5IOU |
|---|---|---|
| Crack1 | 0.2549 | 0.2571 |
| Crack2 | 0.3846 | 0.6436 |
| Joint | 0.2653 | 0.2374 |
| Patching | 0.6393 | 0.5065 |
| Filling | 0.2687 | 0.2888 |
| Pothhole | 0.5385 | 0.5000 |
| Manhole | 0.9756 | 0.7692 |
| Stain | 0.2222 | 0.1667 |
| Shadow | 0.2811 | 0.2877 |
| Marking | 0.5493 | 0.5253 |
| Scratch | 0.5160 | 0.5313 |
| Patching2 | 0.5000 | 0.2941 |

Mask RCNN ResNet 50 Atrous coco

| Index | |
|---|---|
| Precision mAP | 0.2104 |
| Precision mAP@.50IOU | 0.4140 |
| Precision mAP@.75IOU | 0.1934 |
| Precision mAP (large) | 0.2664 |
| Precision mAP (medium) | 0.1571 |
| Precision mAP (small) | 0.0452 |
| Recall AR@1 | 0.1895 |
| Recall AR@10 | 0.3172 |
| Recall AR@100 | 0.3417 |
| Recall AR@100 (large) | 0.3682 |
| Recall AR@100 (medium) | 0.3254 |
| Recall AR@100 (small) | 0.1413 |
| BoxClassifierLoss classification loss | 0.6451 |
| BoxClassifierLoss localization loss | 0.4884 |
| BoxClassifierLoss mask loss | 0.9162 |
| RPNLoss localization loss | 1.5150 |
| RPNLoss objectness loss | 0.3356 |
| Total Loss | 3.9003 |

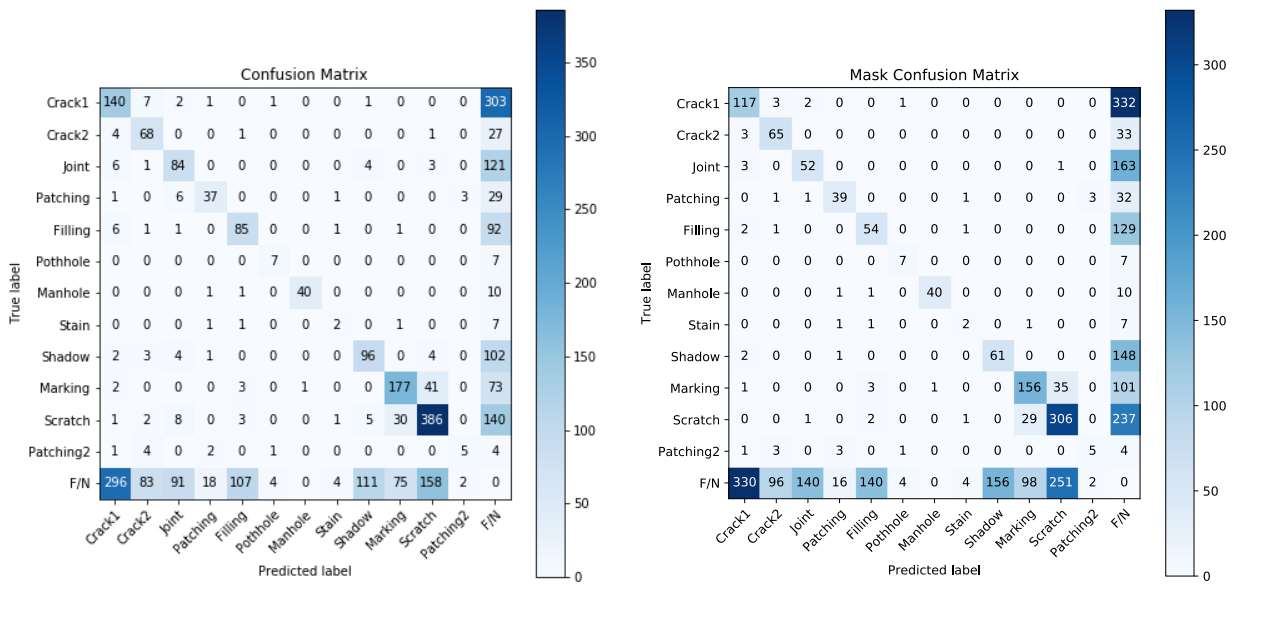

Fig. 6. *Confusion Matrix of Mask RCNN Inception ResNet 50 Atrous COCO on bounding boxes (left) and segmentation masks (right)*

Confusion matrices in Fig. 6, indicate substantial "left detection" (as noted in previous two models) for linear cracks, joints, fillings, shadows, patchings with grid cracks (left detection counts exceed the correctly predicted). The substantial "unclassified detections" counts (which exceed 50% of True Positive) are found in, linear cracks, grid cracks, joints, fillings, potholes, stains, shadows, scratches (only on segmentation masks) and patchings with grid cracks.

*4) Inception ResNet101 Atrous COCO:*

TABLE VIII and TABLE IX show Inception ResNet101 Atrous COCO's mAP(IoU=.50:.05:.95), mAP(IoU=.50) and mAP(IoU=.75), which are 0.2779, 0.4787 and 0.2853 on bounding boxes, and 0.2118, 0.4036 and 0.2045 on segmentation masks respectively, the latter show the highest precisions among those four models. Average Recalls are 0.4975, 0.3759 and 0.1665 on bounding boxes and 0.3404, 0.3382 and 0.1475 on segmentation masks for large, medium and small objects respectively.

TABLE VIII. MODEL METRICS OF MASK R-CNN INCEPTION RESNET101 ATROUS COCO ON BOUNDING BOXES

Precision | Recall

| tag | precision_@0.5IOU | recall_@0.5IOU |
|---|---|---|
| Crack1 | 0.3861 | 0.3055 |
| Crack2 | 0.4962 | 0.6535 |
| Joint | 0.4624 | 0.3653 |
| Patching | 0.6154 | 0.5195 |
| Filling | 0.5195 | 0.4278 |
| Pothhole | 0.5000 | 0.5000 |
| Manhole | 0.9111 | 0.7885 |
| Stain | 0.1538 | 0.1667 |
| Shadow | 0.4889 | 0.4151 |
| Marking | 0.6440 | 0.5421 |
| Scratch | 0.6162 | 0.6858 |
| Patching2 | 0.5000 | 0.2353 |

Mask RCNN ResNet101 Atrous coco

| Index | |
|---|---|
| Precision mAP | 0.2779 |
| Precision mAP@.50IOU | 0.4787 |
| Precision mAP@.75IOU | 0.2853 |
| Precision mAP (large) | 0.3333 |
| Precision mAP (medium) | 0.1979 |
| Precision mAP (small) | 0.0624 |
| Recall AR@1 | 0.2311 |
| Recall AR@10 | 0.4238 |
| Recall AR@100 | 0.4621 |
| Recall AR@100 (large) | 0.4975 |
| Recall AR@100 (medium) | 0.3759 |
| Recall AR@100 (small) | 0.1665 |
| BoxClassifierLoss classification loss | 0.6699 |
| BoxClassifierLoss localization loss | 0.4772 |
| BoxClassifierLoss mask loss | 0.9982 |
| RPNLoss localization loss | 1.5960 |
| RPNLoss objectness loss | 0.4570 |
| Total Loss | 4.1983 |

TABLE IX. MODEL METRICS OF MASK R-CNN INCEPTION RESNET101 ATROUS COCO ON SEGMENTATION MASKS

Precision | Recall

| tag | precision_@0.5IOU | recall_@0.5IOU |
|---|---|---|
| Crack1 | 0.3593 | 0.2835 |
| Crack2 | 0.5038 | 0.6634 |
| Joint | 0.2733 | 0.2146 |
| Patching | 0.6308 | 0.5325 |
| Filling | 0.3182 | 0.2620 |
| Pothhole | 0.5000 | 0.5000 |
| Manhole | 0.9111 | 0.7885 |
| Stain | 0.1538 | 0.1667 |
| Shadow | 0.3143 | 0.2594 |
| Marking | 0.6400 | 0.5387 |
| Scratch | 0.5523 | 0.6146 |
| Patching2 | 0.5000 | 0.2353 |

Mask RCNN ResNet101 Atrous coco

| Index | |
|---|---|
| Precision mAP | 0.2118 |
| Precision mAP@.50IOU | 0.4036 |
| Precision mAP@.75IOU | 0.2045 |
| Precision mAP (large) | 0.2624 |
| Precision mAP (medium) | 0.1618 |
| Precision mAP (small) | 0.0270 |
| Recall AR@1 | 0.1936 |
| Recall AR@10 | 0.3265 |
| Recall AR@100 | 0.3484 |
| Recall AR@100 (large) | 0.3404 |
| Recall AR@100 (medium) | 0.3382 |
| Recall AR@100 (small) | 0.1475 |
| BoxClassifierLoss classification loss | 0.6699 |
| BoxClassifierLoss localization loss | 0.4772 |
| BoxClassifierLoss mask loss | 0.9982 |
| RPNLoss localization loss | 1.5856 |
| RPNLoss objectness loss | 0.4455 |
| Total Loss | 4.1764 |

Fig. 7 confusion matrices of InceptionResNet101 Atorus COCO model indicate substantial "left detection" as noted in the previous model. This is observed on bounding boxes in linear cracks, joints, fillings, stains, shadows, and patchings with grid cracks (left judgement counts exceed the correctly predicted), while patching is more than 50% of True Positive. Substantial "false unclassified detections" counts (which exceed 50% of True Positive) are also found on boundary boxes and segmentation masks in linear cracks, grid cracks, joints, fillings, potholes, stains, shadows, scratches (only segmentation masks) and patchings with grid cracks.

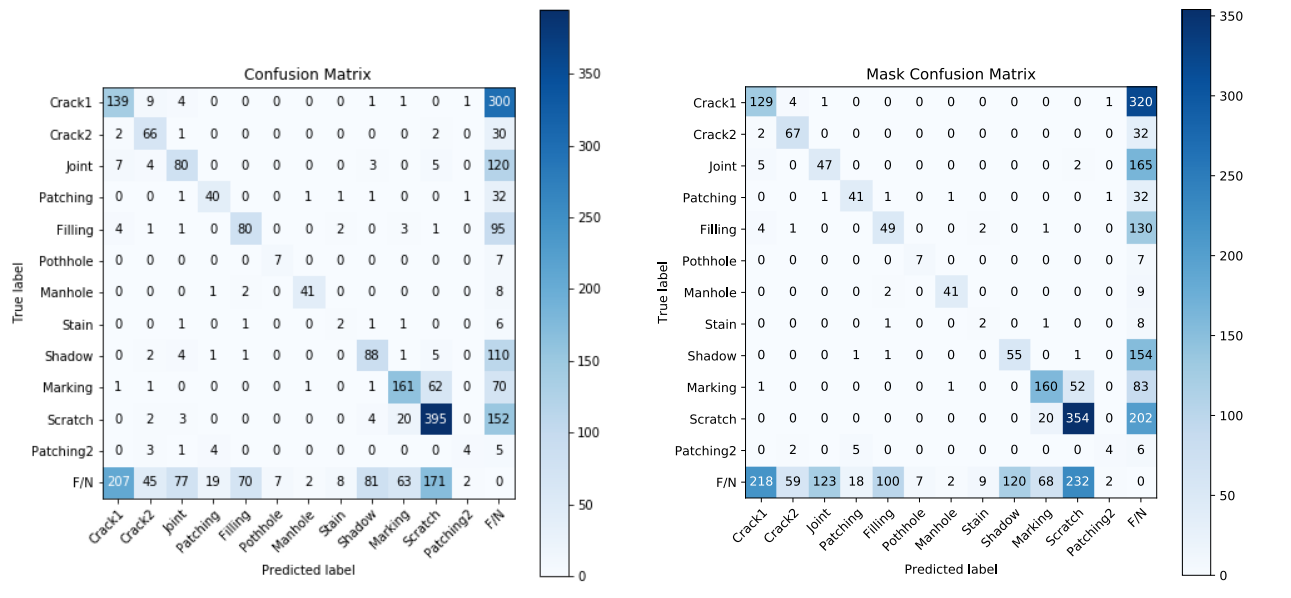

Fig. 7. *Confusion Matrix of Mask RCNN Inception ResNet 100 Atrous COCO on bounding boxes (left) and segmentation masks (right)*

### 5) *Result Summary of TensorFlow Object Detection API Models*

Comparing the results obtained in all four pre-trained Mask R-CNN models, mask R-CNN resnet101 atrous coco (model four) has shown the best precisions and superior mean Average Precisions (small), Average Recalls@100, Average Recalls@10 and Average Recalls@100(small). mAPs generally reached the quick peak, but ARs gradually increased.

From the confusion matrices of all four models, there are substantial number of counts for left detection, i.e. "the model could not fit ground truths with detected objects". This can be seen in the last horizontal row of each confusion matrix, there were substantial number of "unclassified detections" in many classes as well.

## B. *APs in training progresses of four models*

The progress of mAPs of those four models up to 100k steps are shown in Figure 8. The model four showed the advantageous performances in mAP(IoU=.50:.05:.95), and in other five mAP metrics, the faster initial precision increases and the sustaining higher precisions until 100k steps. The model two showed better precision in mAP(IoU=.75) from 50k to 80k steps, than the model four, however, the precision of the model four reached very close to that of model two at 100k steps. The significant higher mAP for small objects (the bottom right graph) of model four is notable. Although the model two reached the higher precision at 100k steps in several metrics, the model four showed better overall performances than other models. On the other hand, the model one showed the least performances in all six AP metrics.

## C. *ARs in training progresses of four models*

All six Average Recall metrics (the six graphs) showed the significant higher recall rate progresses of the model four at most learning steps (Figure 9). The faster performance increases and the stable plateaus in all six AR metrics of the model four are notable. The overwhelming higher recalls of the model four to other models in six AR metrics were prominent compared with mAPs, where some of those mAP metrics were very similar to the model two, as previously noted. A special attention was drawn on much higher AR@100 for the small objects. Although the AR performances of the model one is similar to those in Aps, there is a notable less performance of the model two in AR@100 for the small objects.

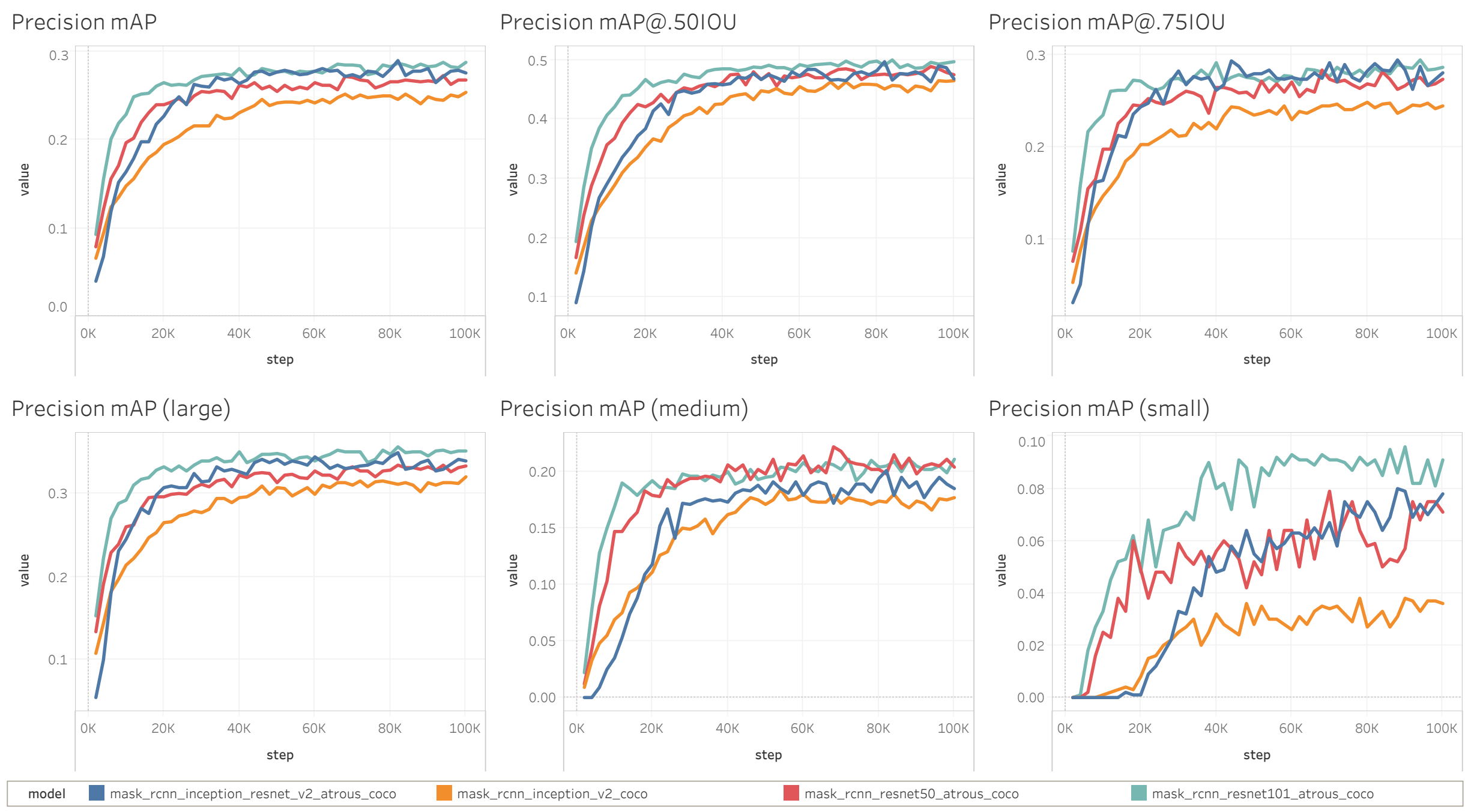

Fig. 8. Mean average precisions (mAPs) of the four model

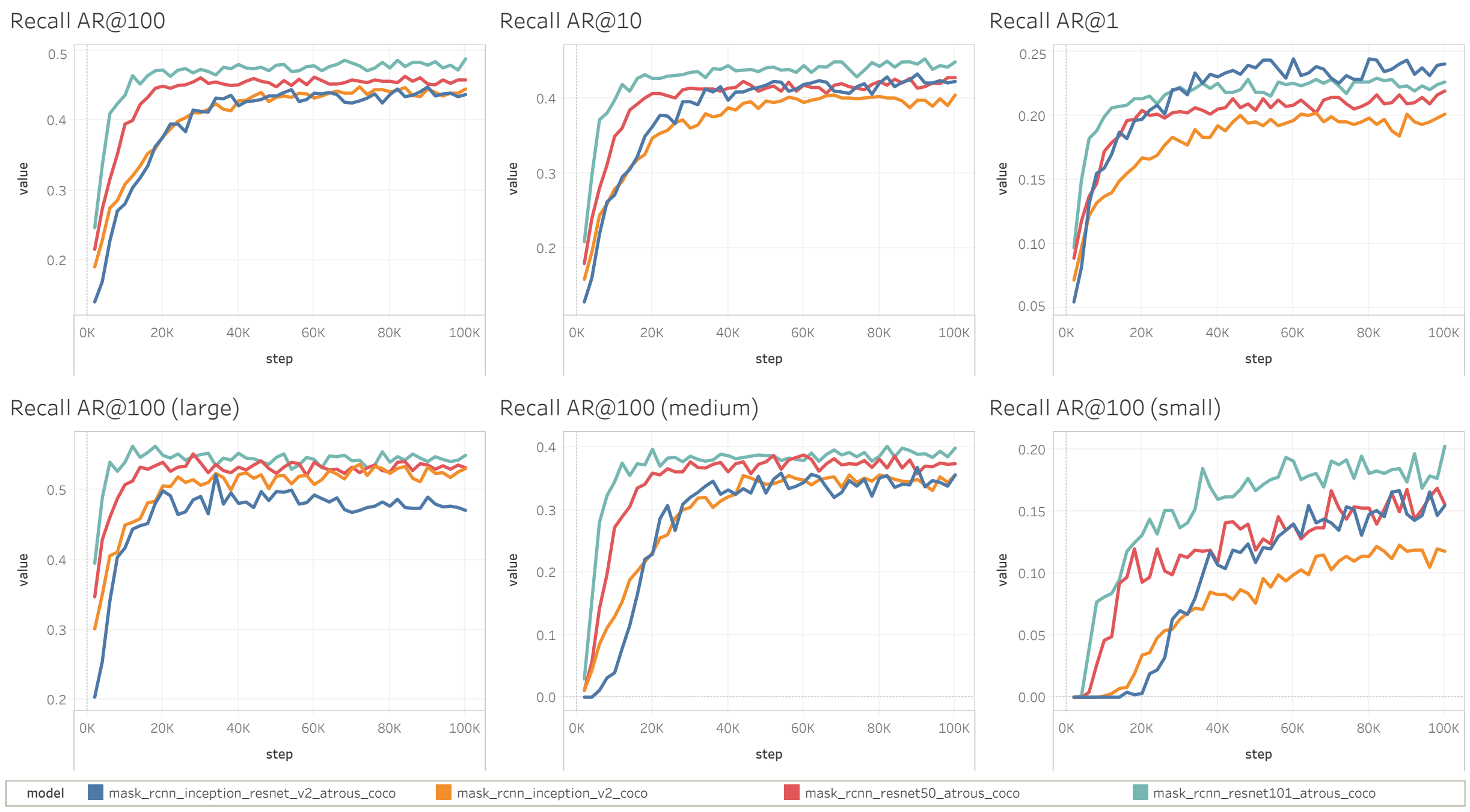

Fig. 9. Average recalls (ARs) of the four models

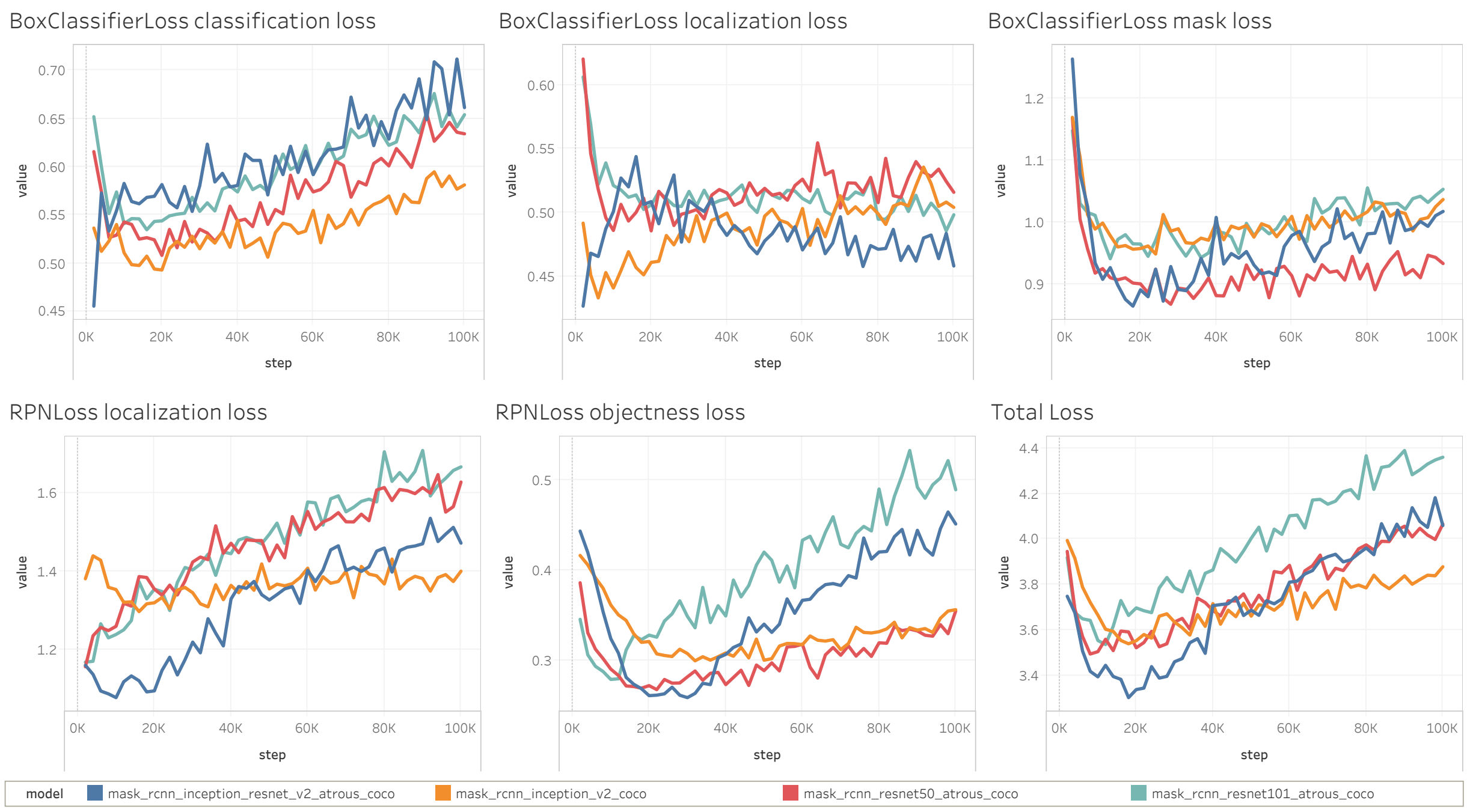

Fig. 10. Loss metrics of the four models

### *D. Losses in training progresses of four models*

Six loss metrics were generated as shown in Figure 10.

There is a generic tendency in all six losses metrics that all four models showed the initial quick drops at 10k-20k followed by the gradual linear increases as steps increase. The gradients of the linear increases vary among those four models. Notable higher total loss of the model four is found.

### *E. Comparison of mAPs/ARs per class*

The AP/AR@0.5IoU of those twelve classes are shown in Figure 11 and 12. The large fluctuations in AP@0.5IoU of pot-holes, stains, and grid cracks in patches are found. After the initial changes, APs@0.5IoU of patchings, fillings, grid cracks, manholes, shadows, markings and scratches are stable. A slight moderate decrease of APs@0.5IoU in linear cracks and joints, and the moderate increase in shadow are found. Compared with those APs@0.5IoU, those metrics of ARs@0.5IoU in all twelve classes are very stable, not much fluctuations as seen in APs@0.5IoU. Except for grid cracks in patchings, all ARs@0.5IoU in other eleven classes are moderately increased.

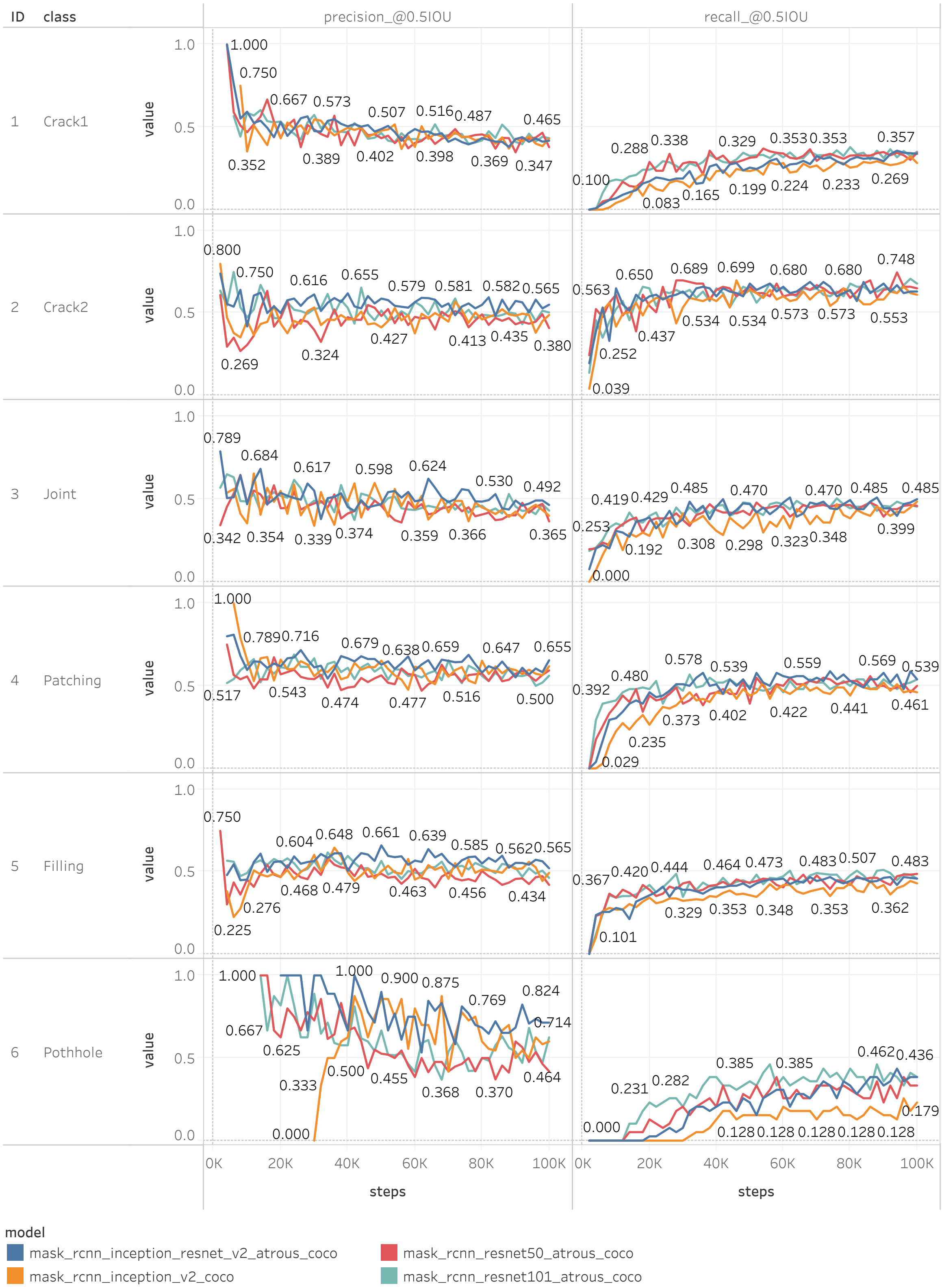

Fig. 11. AP(IoU=.50) and AR(IoU=.50) of twelve road surface object classes (Part 1)

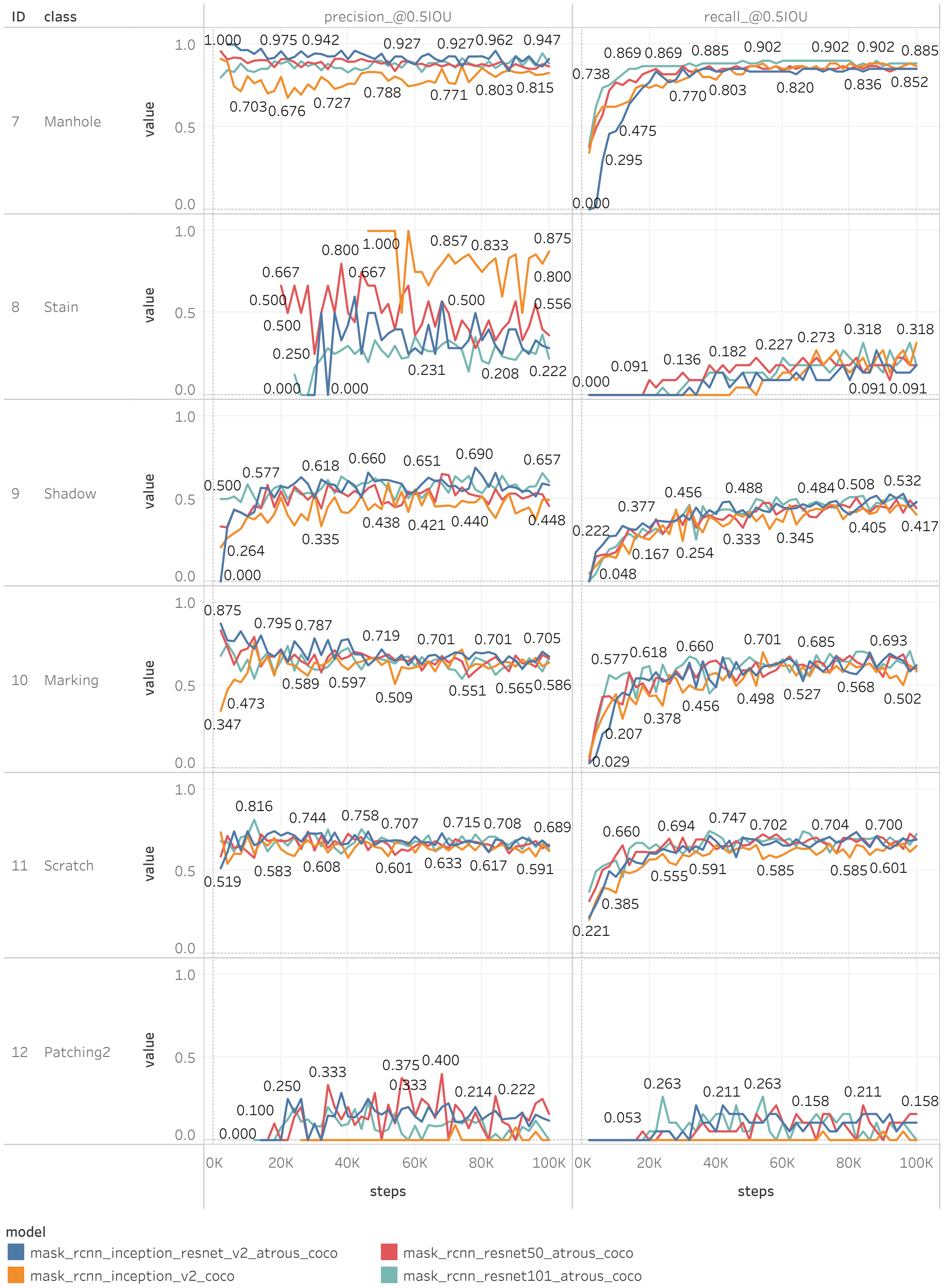

Fig. 12. AP(IoU=.50) and AR(IoU=.50) of twelve road surface object classes (Part 2)

## VI. DISCUSSION

### A. Comparison of four Mask R-CNN models

For easier model identification, they are named as follows:

- Model #1: Mask R-CNN Inception v2 coco;
- Model #2: Mask R-CNN Inception ResNet v2 Atrous coco;
- Model #3: Mask R-CNN Inception ResNet50 Atrous coco;
- Model #4: Mask R-CNN Inception ResNet101 Atrous coco.

It is of interest to describe the performance of the four models before and after the road annotation training. Refer to TABLE X. After 200k labeled image training of COCO datasets, the highest mAPs (IoU=.50:.05:.95) is found in model #2, with value of 0.36. This is followed by model #4 at value 0.33. Model #3 achieves value of 0.29 and the least performing model is model #1 with value 0.25. Substantial differences are observed in terms of detection performance among the four Mask R-CNN models.

TABLE X. MAP(IOU=.50:.05:.95)*100 PERFORMANCE

| Model | Pre-trained model on COCO Dataset | Fine-tuned model with Road Dataset |
|---|---|---|
| mask_rcnn_inception_resnet_v2_atrous_coco | 0.36 | 0.290 |
| mask_rcnn_inception_v2_coco | 0.25 | 0.254 |
| mask_rcnn_resnet50_atrous_coco | 0.29 | 0.268 |
| mask_rcnn_resnet101_atrous_coco | 0.33 | 0.288 |

After relatively small dataset training of those road annotation under 12 classes, the differences in mAP performances are much smaller than those without road object training. The highest was recorded for model #2, but the difference with the second highest model #4 is negligible, where the overall performance is much higher in model #4. The performance reduction in the model #3 and model #1, were also not as much as found in performances without road annotation training.

The detection precision metrics of mAP (IoU=.50:.05:.95) were smaller in those models with road annotation data training, compared with those without training of road annotation data, although it is not adequate for making a direct comparison. However, those objects of the twelve classes on the road surfaces are not as much cognitive as those in the COCO datasets, in human eyes. Therefore, it is reasonable to conclude that the performance of models by road annotation training is not so bad, if one considers the severity of mAP (IoU=.50:.05:.95), which is the primary challenge metric in COCO [13].

The overall model architectures are almost the same in all four models, with a difference in Feature Extractor. Model #1 can be classified as an Inception model, and other three models are combination of Inception and Resnet, where Residual Connection was added on top of Inception architecture [14].

It is notable that strides of Feature Extractor and Anchor Generator of model #1, are both 16, and those of other models are 8. It is also noted that the initial crop size of the Region Proposal Network in model #2 is 17, and those of other models are 14. There is a tendency that smaller parameter size of above modules brings capability to detect smaller objects. Those parameters are the same in model #3 and model #4. However, higher layer numbers in model #4 is advantageous than smaller layer numbers in model #3.

Therefore, it can be summarized that smaller number of strides of Feature Extractor and Anchor Generator, as well as smaller numbers of initial size of the Region Proposal Network, with more layers, made model #4 the best performance with special advantage on small object detection.

In the confusion matrices of all four models, there are substantial number of counts of no class prediction, namely "leaving judgement" or false negatives. In the last horizontal row there are substantial number of counts which have no class definition, although the horizontal row should be the true class, or ground truth. This enigma might be associated with threshold settings which makes direct influences on cut off of boundary boxes counts during the calculation.

### B. IoU prioritized calculation problems of confusion matrix

Confusion matrix is a cross table of class detection counts in column, and class ground truth counts in rows. The diagonal cells of each corresponding class are true positives, i.e., correctly detected and classified instances. The bottom end of each column is counts of false positives, and right end of each row is counts of false negatives.

However, there is a fundamental problem of confusion matrix for regional object detection tasks, of which models are supposed to detect both a correct region and a correct object class simultaneously. As regional detection always sets an IoU (Interception over Unit) threshold of overlap extent of detected and ground truth areas, both in the bounding boxes and the segmentation masks. Therefore the "correct detection" varies upon threshold variable. This study adopts IoU=0.5 as both thresholds for AP and AR metrics according to the COCO detection standard.

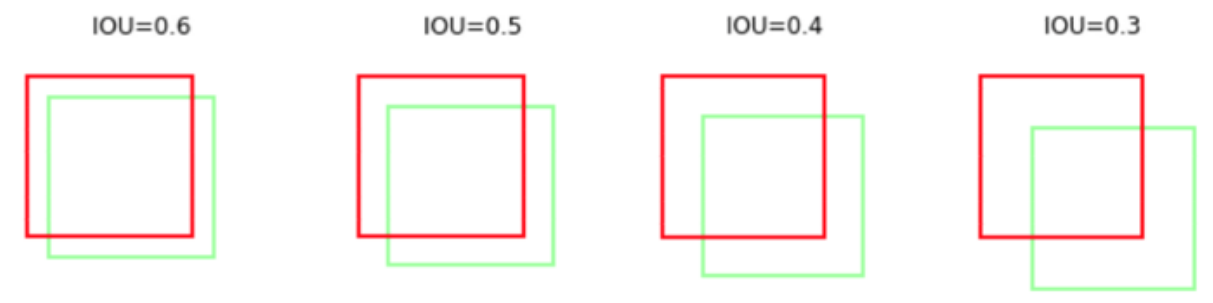

Fig. 13. Examples of various IoUs (0.3~0.6)

Refer Fig. 12 for IoU examples. The red color is a ground truth area, and the green color is a detected area at each IoU level. The COCO standard's IoU=0.5 requires a substantial overlap extent in ground truth and detected areas.

However, some of regional object detection tasks do not require such a high standard, where an overlap extent of IoU=0.3 or IoU=0.2 might be quite enough, hence, the "regional correct detection" does not have a uniform standard.

The most commonly used regional detection algorithm created by Santiago Valdarrama, prioritizing the IoU conducts the following steps [15, 16].

1 calculates IoU with ground truths and detected areas, and makes a list of over threshold detection candidates

2 lists in IoU decent order

3 if there are several detection candidates over a single ground truth, discard all candidates by leaving one candidate with a maximum IoU

4 lists the remaining in decent order

5 if a single detection candidate matches multiple ground truths, discard all candidates by leaving one candidate with a maximum IoU

6 check the ground truth one by one

7 if a ground truth matches any detected candidates, count up it in a confusion matrix cell of a corresponding ground truth and detection class (true positive and false positive)

8 if a ground truth does not match any detected candidates, count up it in the right end cell of corresponding ground truth class (false negative)

9 check the detection candidates one by one

10 if one candidate does not match any ground truth, count up it in the bottom row of a corresponding detection class (false positive)

Our modified confusion matrix algorithm created by Masatoshi Itagaki is as followed.

1 Sort detected candidates in descend confidence score order

2 Pick one ground truth, if it has previously matched detected candidates pick next ground truth

3 Pick one detected candidate and calculate IoU

4 If a ground truth matches with detected candidates of the same class, select one with highest IoU as a match

5 If a ground truth has no detected candidates with the same class, select one of other classes with highest IoU as a "False Positive" match

6 If a detected candidate matches with more than a single ground truth instance, select a ground truth of higher IoU prioritizing the class

7 Repeat 2 to 7 until no more changes take place

8 Matches are counted up in corresponding class ID cells of the confusion matrix (as true positives and false positives)

9 If a ground truth has no matching detected candidates, count up right end column cells of corresponding ground truth class as false negatives

10 If a detected candidate has no ground truths matched, count up at the bottom row cells of corresponding class as false positives

The comparison of conventional and modified calculation methods is made at thresholds of IoU=0.5 and Confidence Score=0.5, for bounding boxes, and segmentation masks. The results are shown as in Figure 14 and Figure 15.

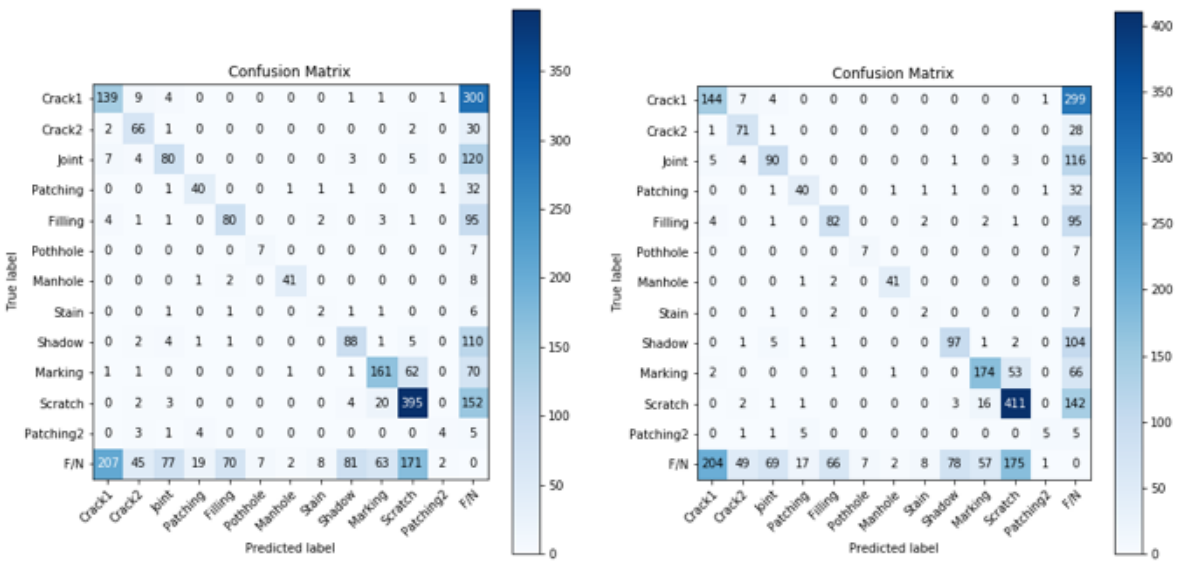

Fig. 14. Conventional (left) and modified (right) confusion matrices of bounding boxes

TABLE XI. CLASS METRICS BY CONVENTIONAL(LEFT), MODIFIED(RIGHT) CONFUSION MATRICES OF BOUNDING BOXES

| category | precision_@0.5IOU | recall_@0.5IOU | category | precision_@0.5IOU | recall_@0.5IOU |
|---|---|---|---|---|---|
| Crack1 | 0.3861 | 0.3055 | Crack1 | 0.4000 | 0.3165 |
| Crack2 | 0.4962 | 0.6535 | Crack2 | 0.5259 | 0.7030 |
| Joint | 0.4624 | 0.3653 | Joint | 0.5172 | 0.4110 |
| Patching | 0.6154 | 0.5195 | Patching | 0.6154 | 0.5195 |
| Filling | 0.5195 | 0.4278 | Filling | 0.5325 | 0.4385 |
| Pothhole | 0.5000 | 0.5000 | Pothhole | 0.5000 | 0.5000 |
| Manhole | 0.9111 | 0.7885 | Manhole | 0.9111 | 0.7885 |
| Stain | 0.1538 | 0.1667 | Stain | 0.1538 | 0.1667 |
| Shadow | 0.4889 | 0.4151 | Shadow | 0.5389 | 0.4575 |
| Marking | 0.6440 | 0.5421 | Marking | 0.6960 | 0.5859 |
| Scratch | 0.6162 | 0.6858 | Scratch | 0.6372 | 0.7135 |
| Patching2 | 0.5000 | 0.2353 | Patching2 | 0.6250 | 0.2941 |

The true positive counts increased in most classes shown as in the diagonal cells of the modification (right graph) is notable. The slight decreases are found in false positives (the remaining counts from the true positive and the right end column counts). The false negative counts at the right end column, which are those left over ground truth counts from detection at those thresholds, shows few count changes, except for the scratch class, of ten counts reduction. The modified calculation method seems to reducing unnecessary rejection of true positives and false positives.

Refer Table XI. The notable increases in both precision and recall except for patching, pothole, manholes and stains classes.

The same comparisons of the conventional and the modified methods for segmentation masks in confusion matrices and class metrics are also made as in Figure 15. and Table XII. The results show almost no changes between the calculation methods, both in the confusion matrices and the class metrics. Very slight increases in metrics are found in grid crack, filling, marking and scratch classes but very slight decreases in linear crack and shadow classes.

In the modified calculation algorithm, if a ground truth does not match detection candidates with less class threshold IoU, but above threshold IoU in another class, it will be counted as a false negative. The less performance on segmentation masks by the modified calculation might be associated with smaller sizes of segments than bounding boxes, which leads less cases of above false negative counts.

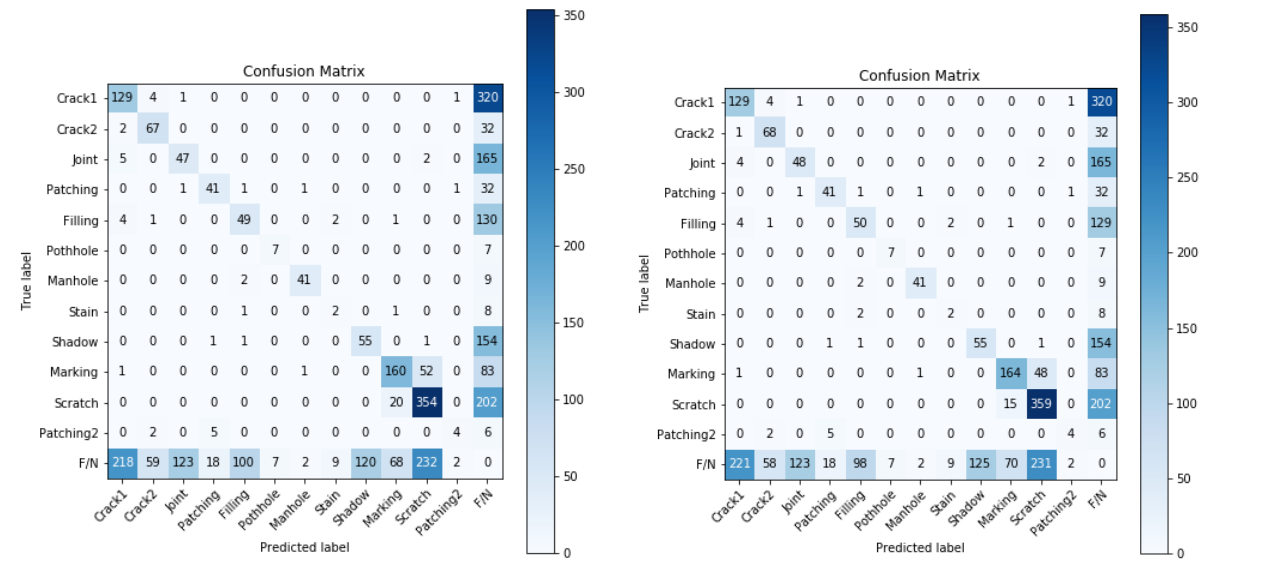

Fig. 15. Conventional (left) and modified (right) confusion matrices of segmentation masks

TABLE XII. CLASS METRICS BY CONVENTIONAL(LEFT), MODIFIED(RIGHT) CONFUSION MATRICES OF SEGMENTATION MASKS

| category | precision_@0.5IOU | recall_@0.5IOU | category | precision_@0.5IOU | recall_@0.5IOU |
|---|---|---|---|---|---|
| Crack1 | 0.3593 | 0.2835 | Crack1 | 0.3583 | 0.2835 |
| Crack2 | 0.5038 | 0.6634 | Crack2 | 0.5113 | 0.6733 |
| Joint | 0.2733 | 0.2146 | Joint | 0.2775 | 0.2192 |
| Patching | 0.6308 | 0.5325 | Patching | 0.6308 | 0.5325 |
| Filling | 0.3182 | 0.2620 | Filling | 0.3247 | 0.2674 |
| Pothhole | 0.5000 | 0.5000 | Pothhole | 0.5000 | 0.5000 |
| Manhole | 0.9111 | 0.7885 | Manhole | 0.9111 | 0.7885 |
| Stain | 0.1538 | 0.1667 | Stain | 0.1538 | 0.1667 |
| Shadow | 0.3143 | 0.2594 | Shadow | 0.3056 | 0.2594 |
| Marking | 0.6400 | 0.5387 | Marking | 0.6560 | 0.5522 |
| Scratch | 0.5523 | 0.6146 | Scratch | 0.5601 | 0.6233 |
| Patching2 | 0.5000 | 0.2353 | Patching2 | 0.5000 | 0.2353 |

## VII. CONCLUSION

This paper reports the final evaluation results obtained by test datasets and modified confusion matrix methodologies.

There are substantial number of counts of left detection, as well as unclassified detections.

A modified confusion matrix model to avoid prioritizing IoU is tested. There are notable true positive increases in bounding box detection, but almost no changes found in segmentation masks.

The less performance on segmentation masks by the modified calculation might be associated with smaller sizes of segments than bounding boxes. This necessitates a further research.

## ACKNOWLEDGMENT

Authors owe to Niigata University of International and Information Studies for funding support on the "Mobile Sensing Development by Deep Learning" project in 2020.

Special thanks are due to Fukuda Road Co. Ltd for its provision of road images for the annotation works as well as for the provision of information on Multi Fine Eye.

Finally Authors owe much to Tsuyoshi Ikenori for his works on the annotation data.